\documentclass[10pt,journal,twoside]{IEEEtran}

\usepackage[pdftex]{graphicx}
\usepackage{cite}

\usepackage{amsmath,amssymb,amsfonts,amsthm}

\usepackage{url}

\usepackage{cite}
\usepackage{graphicx}
\usepackage{multicol,multirow}
\usepackage[usenames,dvipsnames]{xcolor}
\usepackage{textcomp}
\usepackage{adjustbox}
\usepackage{diagbox}
\usepackage{booktabs}
\usepackage{lettrine}
\usepackage{siunitx}
\usepackage{caption,subcaption}
\usepackage{tcolorbox}
\usepackage{hyperref}
\usepackage{algorithm}
\usepackage[noend]{algpseudocode}
\usepackage{txfonts}
\usepackage{soul, xcolor}

\let\oldproofname=\proofname
\renewcommand{\proofname}{\rm\bf{\oldproofname}}

\newsavebox{\twosubbox}
\begin{document}

\title{Anticipating Degradation: A Predictive Approach to Fault Tolerance in Robot Swarms}

 \author{James~O’Keeffe, \textit{Member, IEEE}%

 \thanks{Manuscript received: April 9, 2025; Revised:
June 18, 2025; Accepted: July 8, 2025. This paper was recommended for publication by Editor Cosimo Della Santina upon evaluation of the Associate Editor and Reviewers’ comments. The work of J. O'Keeffe was supported by the Royal Academy of Engineering UK IC Postrdoctoral Fellowship award under Grant ICRF2223-6-121. J. O'Keeffe is with the Department of Computer Science, University of York, United Kingdom. {\tt\small james.okeeffe@york.ac.uk}. Digital Object Identifier (DOI): see top of this page.}%

 }

\IEEEtitleabstractindextext{%
\begin{abstract}
An active approach to fault tolerance is essential for robot swarms to achieve long-term autonomy. Previous efforts have focused on responding to spontaneous electro-mechanical faults and failures. However, many faults occur gradually over time. This work argues that the principles of predictive maintenance, in which potential faults are resolved before they hinder the operation of the swarm, offer a promising means of achieving long-term fault tolerance. This is a novel approach to swarm fault tolerance, which is shown to give a comparable or improved performance when tested against a reactive approach in almost all cases tested.
\end{abstract} 

}

\maketitle
\markboth{IEEE ROBOTICS AND AUTOMATION LETTERS. PREPRINT VERSION. ACCEPTED JULY, 2025}{O'Keeffe: Anticipating Degradation: A Predictive Approach to Fault Tolerance in Robot Swarms}
\IEEEdisplaynontitleabstractindextext
\IEEEpeerreviewmaketitle
\vspace{1cm}
\IEEEraisesectionheading{\section{Introduction \& Related Work}\label{sec:introduction}}
\IEEEPARstart{T}{}here is increasing interest in the application of Swarm Robotic Systems (SRS) to tasks across a wide variety of sectors and scenarios \cite{schranz2020swarm}. However, a significant barrier to the deployment of autonomous robots in many real-world applications is the risk of failure or loss of autonomous control in the field. SRS enjoy a degree of innate robustness -- the ability to tolerate faults and failures in individual robots without significant detriment to the swarm as a whole -- because of their redundancy and distributed control architectures \cite{csahin2004swarm}. While this is true in some cases, previous research shows that partially failed robots that exert influence on other robots in the swarm can significantly degrade overall performance \cite{winfield2006safety}, concluding that an active approach to fault tolerance is necessary if SRS are to retain long-term autonomy \cite{bjerknes2013fault}. 

Active fault tolerance comprises some combination of autonomous fault detection, diagnosis, and recovery (FDDR) \cite{o2018fault}. Most previous work towards fault tolerance in SRS has focused on injecting sudden sensor and actuator faults into individual robots, as if they had failed spontaneously (e.g. \cite{tarapore2019fault,o2018fault,khadidos2015exogenous,carminati2024distributed}).

Once a robot is detected as faulty, an action must be taken to resolve the fault or otherwise prevent further detriment to system performance. So far, efforts towards fault tolerant SRS have focussed on reactive approaches to fault resolution -- i.e. resolving faults \textit{after} they have manifested as failures. These approaches usually adopt one of the following strategies/assumptions: robots can autonomously repair themselves and/or other robots in the field \cite{oladiran2019fault,o2018fault} or; a failed robot can simply be shut down and abandoned \cite{khadidos2015exogenous}, prevented from interfering \cite{strobel2023robot}, or its disruptive potential neutralised by mitigating actions taken by the swarm \cite{bossens2021rapidly}. Other potential options include: robots can be repaired by a human in the field or; failed robots can be retrieved autonomously or by a human and brought to a site where they can be repaired. However, there are conditions and limitations to each of the described approaches.

Robotic platforms available at the time of writing are broadly unable to autonomously self-repair or repair other failed robots in the field. Such functionality requires additional components, actuation, and control for each individual robot that could impose unaffordable costs to many SRS applications. Failed robots could be repaired or retrieved by a human in safe, open, controlled environments, but this may not be possible in environments that are inaccessible or dangerous, the latter of which is highlighted among the motivating use-case scenarios for SRS \cite{csahin2004swarm}. Similarly, autonomous retrieval of failed robots, either with appropriate manipulating actuators or via coordinated `shunting', may be feasible in some scenarios, but is itself a challenging control problem that likewise may not be available to all platforms or possible in inaccessible or dangerous environments.

\begin{figure*}[!t]
    \includegraphics[width=\textwidth]{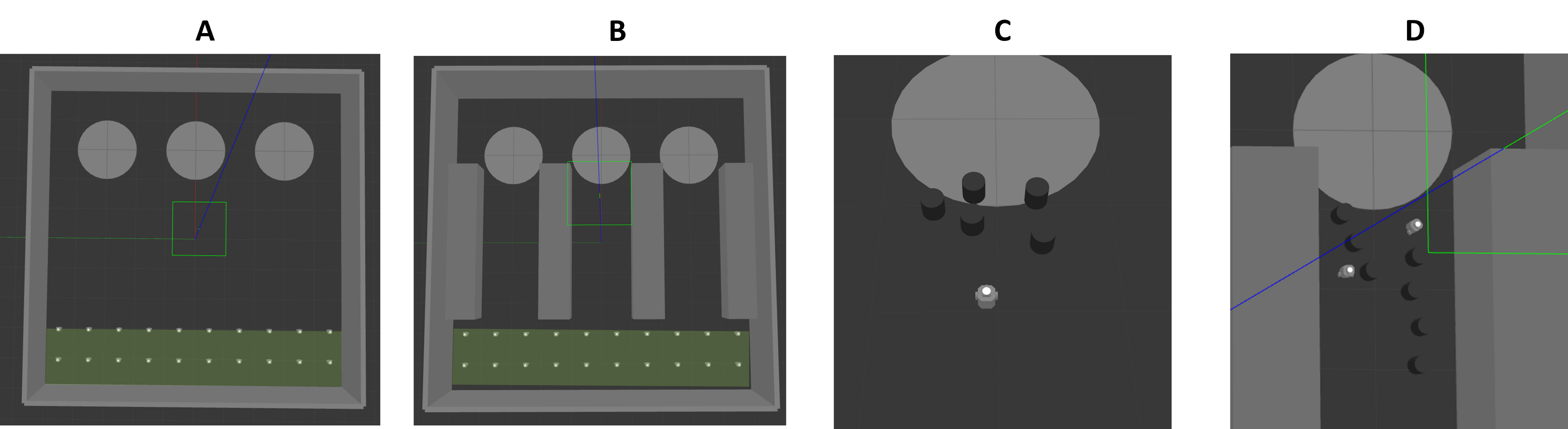}
    \caption{\textbf{A:} Experimental setup for 20 robots in the open environment. The base is highlighted in light green and resource nests are indicated by the three grey circles opposite the robots. \textbf{B:} Experimental setup for 20 robots in the constrained environment. \textbf{C:} Example of how clusters of shutdown faulty robots can impede swarm progress by obstructing operational robots from their goals. For ease of user differentiation during experiments, shutdown robots appear as dark featureless cylinders of equivalent dimensions to the lighter coloured functioning robots. \textbf{D:} Example of severe disruption caused by robots shutdown in already constrained spaces, completely blocking access in some cases.} 
 \label{fig:setup}
\end{figure*}

Shutting down, isolating, and/or abandoning faulty robots in the field offers a currently feasible recovery strategy for swarms that solves the swarm anchoring problem highlighted in collective photo-taxis scenarios \cite{winfield2006safety}, and has been adopted in other approaches to swarm fault tolerance (\cite{khadidos2015exogenous,strobel2023robot}). Robots can also modify their own behaviour to mitigate potential disruption caused by a failed robot in some cases \cite{bossens2021rapidly}. However, neither approach explicitly resolves the underlying problem which, if left unchecked, can become problematic in applications requiring a minimum number of functioning robots to operate for extended periods of time.

Although there are scenarios in which modelling faults and failures as spontaneous events is appropriate (e.g. a robot being suddenly immobilised after becoming stuck on an obstacle), previous studies reveal that one of the most common causes of failure is the gradual degradation of sensor and actuator hardware \cite{CarlsonMTBF}. These types of fault have so far been omitted from fault tolerant SRS literature, much of which focuses on identifying sudden failures in a sub-population of robots while the non-faulty robots retain uniform normal functionality. This narrows the problem space, such that the aim of fault detection is simply to detect failures as soon after occurrence as possible. In fact, Carlson et. al. \cite{CarlsonMTBF} highlight that robots have a mean time between failures (MTBF) that indicates the average length of uninterrupted time in which a robot can reliably operate before suffering some kind of fault or failure. If all robots have a MTBF, all robots can be considered as occupying some variable position on a scale of degradation at any point in time. This complicates the differentiation of faulty and normal behaviours. One must ascertain the point at which hardware degradation becomes problematic for the afflicted robot and for the rest of the swarm -- i.e. there must be a defined level of operational acceptability, above which a robot can be considered healthy, and below which a robot can be considered faulty. Determining this level, and thus the desired point of fault detection, is a trade off -- it is inefficient to allow a robot's performance to degrade excessively before attempting to resolve it, just as it is to declare robots faulty for the slightest reductions in performance. 

Examining gradually occurring faults provides opportunities for new approaches to fault resolution and swarm fault tolerance in general. If robots are routinely serviced and repaired within their MTBF, and assuming there are no other adversarial factors, there should be few instances of failure during operation (if any). This is the underlying principle of preventative maintenance -- identifying and resolving potential problems before they cause system downtime or serious damage to hardware. Preventative maintenance is widely employed across industrial machines for its long-term cost effectiveness and improvements to safety and reliability \cite{malik1979reliable}. Similar in concept is \textit{predictive maintenance}. Whereas preventative maintenance schedules maintenance work at regular specified intervals, predictive maintenance aims to schedule the work only as required -- reducing individual downtime and resource expense \cite{zonta2020predictive}. A predictive maintenance approach to swarm fault tolerance, whereby the swarm detects faults early enough to allow the faulty robot a grace period in which to return itself to a safe area for receiving maintenance, could be effective in preventing failure in the field. 

This study proposes a novel SRS method to model robot faults as gradual degradation and demonstrates for the first time that traditional reactive approaches to swarm tolerance are unsustainable in some scenarios if robots cannot be physically removed. This study shows that a predictive maintenance approach is able to offer a competitive or improved performance in almost all cases tested -- verified using an ideal fault detection mechanism as well as a novel fault detection algorithm. Overall, this study presents and validates a new approach to swarm fault tolerance that runs counter to traditionally held views that robot swarms, by their definition, \textit{must} be tolerant to the loss of individual robots \cite{csahin2004swarm}.

The remainder of the paper is structured as follows. Section 2 details experimental implementation and test scenario. Section 3 describes experimental results alongside relevant discussion. Section 4 concludes and identifies areas for future research.

\newcommand{\algorithmautorefname}{Algorithm}
\section{Methodology}
All experiments were conducted with Robot Operating System (ROS) 2 and Gazebo Classic. For ease of reading, a non-exhaustive list is provided in \autoref{table:vars} that gives the definitions for key variables in this work.

\begin{table}
\begin{center}
\begin{tabular}{ |p{0.8cm}|p{5cm}|p{1.3cm}|}
\hline
Symbol & Meaning & Value \\
\hline
$\Delta P$ & Rate of power consumption & \autoref{equ:P}\\
$v$ & Linear velocity & \autoref{equ:velocity}\\
$r$ & Sensing range & \autoref{equ:range_2} \\
$\gamma$ & Variable for monitoring $r$ & \\
$d_{l,r,S}$ & Degradation coefficients (left motor, right motor, sensor) &   \\
$X$ & Current robot behavioural signature repertoire  & \\
$Y$ & Learned behavioural signature repertoire & \\
$p$ & Behavioural signature in $X$ or $Y$ & $|p| = 30$ \\
$m$ & Matching specificity & \autoref{equ:ok_spec} \\
$x_i$ & Corresponding population score for $p_i$ in $X$ & \autoref{equ:ok} \\
$W$ & The most recent entries of robot behavioural data & $|W| = 300$ \\
$d_0$ & The threshold for which a robot with $d_{l,r,S} < d_0$ is considered faulty & \\
$\delta$ & The value of a robot's $d_{l,r}$, whichever is smallest, at the moment a motor fault is detected by \autoref{algAAPD}, or the value of $d_S$ when a sensor fault is detected. & \\

\hline

\end{tabular}
\end{center}
\caption{The parameter selections for \autoref{equ:ok} and \autoref{equ:ok_spec} used by different stages of \autoref{algAAPD}.}
\label{table:vars}
\end{table}
This study considers a robot foraging scenario (a classic benchmark in SRS research \cite{bayindir2016review}), in which robots gather resources in an enclosed arena of 10m x 10m. Robots begin each experiment at an area of the arena defined as the `robot base', which spans the width of the arena along the row $y = 2$ (see \autoref{fig:setup}). The robot base is the area that foraged resources must be returned to and is assumed to be the only part of the arena that can be accessed by non-swarm actors (e.g. human operators or other autonomous non-swarm agents, such as robotic arms). 

A homogeneous SRS of simulated TurtleBot3's, two-wheeled differential drive robots \cite{TB3} with maximum linear velocity $v_{max} = 0.22 ms^{-1}$, are studied. Robots are equipped with sensors to localise obstacles and robots up to a maximum range of $r_{max} = 4m$. Robot locomotion, sensing and communication processes consume power. The rate of power consumption per second of simulated time, $\Delta P$, at any given moment is given by \autoref{equ:P}. 

\begin{equation}
    \Delta P = \Delta P_l + \Delta P_r + \Delta P_S 
    \label{equ:P}
\end{equation}
Where $\Delta P_{l,r}$ is the power consumed by left and right motors, respectively, and $\Delta P_S$ is the power consumed by sensing and communication processes. The maximum rate of power consumption per second of simulated time, $\Delta P_{max}$, is described by \autoref{equ:P_max}. 

\begin{equation}
    \Delta P_{max} = \Delta P_{l_{max}} + \Delta P_{r_{max}} + \Delta P_S = \frac{P_0}{300} 
    \label{equ:P_max}
\end{equation}

Where $P_0$ is a robot's total power capacity ($P_0 = 1$, unitless), $\Delta P_{l_{max},r_{max}} = \frac{2}{5} \Delta P_{max}$, and $\Delta P_S = \frac{1}{5} \Delta P_{max}$. A robot can thus be in the state $\Delta P = \Delta P_{max}$ for 5 minutes of simulated time.

Two types of environments are considered. The `empty' environment consists of the 10m x 10m enclosed arena that is empty apart from 3 resource nests of 1m radius at arena (x,y) coordinates (2,8), (5,8), and (8,8). The `constrained' environment also contains 3 resource nests at the same positions, but the area between the resource nests and the robot base is separated into 3 equally spaced corridors of 2m width and 5m length. Each environment can be seen in \autoref{fig:setup}A-B. 

Two foraging algorithms are considered. The Global Positioning Foraging (GPF) algorithm, described by \autoref{algA}, is a basic foraging algorithm in which each robot performs a random walk exploration until a resource nest comes within sensing range, $r_S$. The robot will then approach the nearest nest, collect a resource, and return to base by the shortest Euclidean path, avoiding any obstacles along the way. The robot has \textit{a priori} knowledge of the location of the base and of itself in a global coordinate frame but not of the location of resource nests, which must be sensed locally. The Local Positioning Foraging (LPF) algorithm, described by \autoref{algB}, is similar, except that robots are not assumed to have access to GPS information. In order to localise, the swarm must form an ad-hoc network. Each robot has a limited localising range, and its status is determined by whether or not there exists a path from a robot to the base that is valid for the variable sensing ranges of each node. A robot is only able to extend a communication chain if it is within the sensing range of the previous node and has line of sight to it. A robot will not move if it cannot localise. 


\begin{algorithm} [t]
\caption{Global Positioning Foraging (GPF) Algorithm}\label{algA}
\begin{algorithmic}[1]
\While {Running}
\If {Object Distance $\leq 0.5m$} avoid
\ElsIf{Resource collected} Return to base
\If{Robot at base} Deposit resource
\EndIf
\ElsIf {Distance to nearest Resource Nest $\leq 0.5m$} Collect resource
\ElsIf {Distance to nearest Resource Nest $\leq r_s$} Approach nearest resource
\Else{} Randomly Explore
\EndIf
\EndWhile
\end{algorithmic}
\end{algorithm}

\begin{algorithm} [t]
\caption{Local Positioning Foraging (LPF) Algorithm}\label{algB}
\begin{algorithmic}[1]
\While {Running}
\If{Distance to closest networked node $\leq 3m$}
\If {Object Distance $\leq 0.5m$} avoid
\ElsIf{Resource collected} Return to base
\If{Robot at base} Deposit resource
\EndIf
\ElsIf {Distance to nearest Resource Nest $\leq 0.5m$} Collect resource
\ElsIf {Distance to nearest Resource Nest $\leq r_s$} Approach nearest resource
\Else{} Random Explore
\EndIf
\Else{} Wait
\EndIf
\EndWhile
\end{algorithmic}
\end{algorithm}

\subsection*{Fault Modelling}
Focus is given to faults occurring by gradual degradation on motor and sensor hardware, e.g. those caused by the build up of dirt and debris \cite{CarlsonMTBF}. The level of degradation on a robot's left motor, right motor, and sensor hardware is indicated by coefficients $d_l$, $d_r$, and $d_S$, respectively.

The power consumed by motors will be affected by their condition. A robot in perfect conditions (i.e., $d_{l,r} = 1$) is taken to cause its motors to operate at 75\% load \cite{ranges26determining}. The effects of motor degradation on power consumption and output linear velocity are then described by \autoref{equ:power_wheels} and \autoref{equ:velocity}, respectively.

\begin{equation}
     \Delta P_{l,r} = \frac{\Delta P_{{l_{max}},{r_{max}}}}{1+e^{-10((1-d_{l,r}) + 0.11)}}    
     \label{equ:power_wheels}
\end{equation}

\begin{equation}
    v_{l,r} = \frac{v_{max}}{1+e^{-5(2d_{l,r} - 1)}} 
    \label{equ:velocity}
\end{equation}

 \begin{figure}[!htb]
    \includegraphics[width=0.48\textwidth]{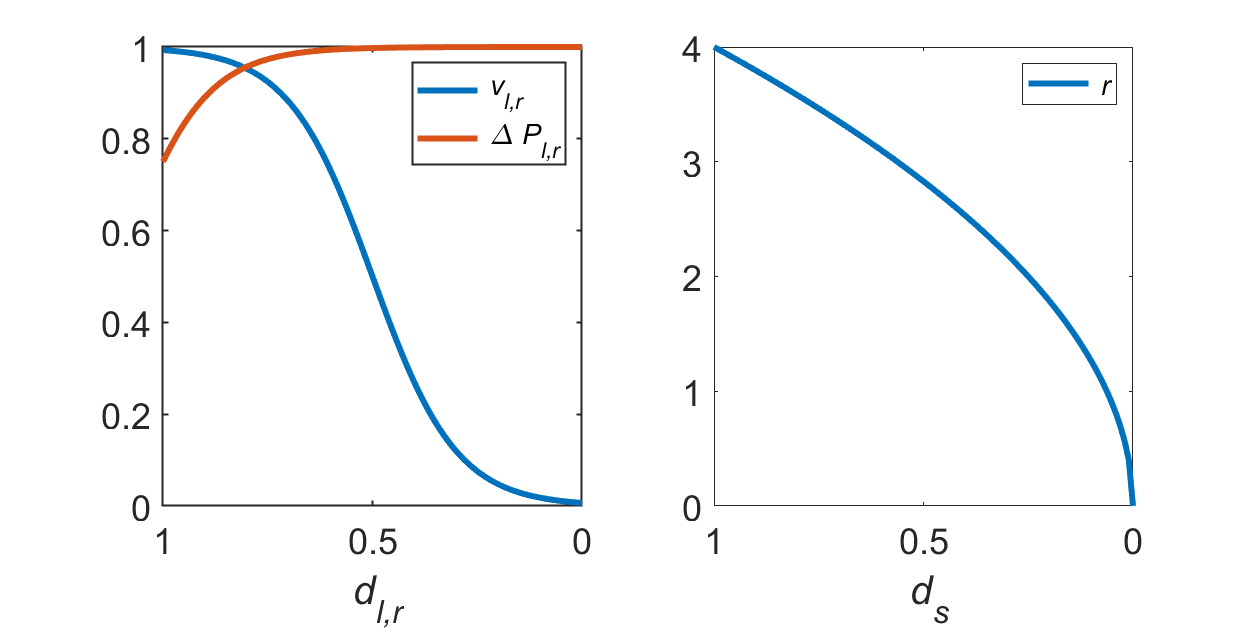}
    \caption{Plots of \autoref{equ:power_wheels} and \autoref{equ:velocity} (left) and \autoref{equ:range_2} (right)} 
    \label{fig:sens_deg}  
\end{figure}

Where values of constants are set to give the intersection of $v_{l,r}$ and $\Delta P_{l,r}$ as shown in the left plot of \autoref{fig:sens_deg}. This intersection reflects that, as the value $d_{l,r}$ increases, motors can initially draw more power to achieve $v_{max}$ but, eventually, the mechanical power required will become greater than that which can be supplied. At this point, degradation will begin to reduce a robots maximum achievable velocity.

The level of degradation affecting a robot's sensor is given by coefficient $d_S$. Since sensor output is not governed by corrective feedback loops in this case, $\Delta P_S$ does not vary with $d_S$, which only affects robot sensing range, $r$. This is modelled according to the inverse square law (\autoref{equ:range_2}) and shown in the right plot of \autoref{fig:sens_deg}.

\begin{equation}
    r = r_{max} \sqrt{d_s} 
    \label{equ:range_2}
\end{equation}

\subsection*{Fault Detection}

Since examining faults by gradual degradation is novel in SRS, existing swarm fault detection models cannot be readily implemented in this scenario. Therefore, a novel distributed fault detection model, based on models of the natural immune system \cite{farmer1986immune}, is provided.

Each robot produces short behavioural signatures consisting of 30 data points sampled at 6Hz. Behavioural signatures are stored in separate independent repertoires, $X_M$ and $X_S$, by each robot according to whether the signature relates to motor or sensor hardware, respectively. Behavioural signatures relating to motor hardware are 3-dimensional, consisting of linear velocity, $v$, angular velocity, $\omega$, and rate of power consumption, $\Delta P$, all normalised to values between 0 and 1. Behavioural signatures relating to robot sensors are 1-dimensional, consisting of a specifically created variable, $\gamma$, which indicates the closest distance at which a given robot $R_1$ can localise neighbour $R_2$, but where $R_2$ is simultaneously unable to localise $R_1$ ($\gamma = r_{max}$ if a robot can mutually localise with each neighbour). Behavioural signatures in $X_M$ and $X_S$ are handled separately by the model and do not interact.

Each robot produces a new signature $p$ after every 5 seconds of simulated time, which is assigned a population value, $x$, initialised to zero. A new behavioural signature, $p_i$, is only added to repertoire $X_M$ or $X_S$ if it does not already contain a similar signature $p_j$ such that $m(p_i,p_j) > 1.5$. The matching specificity, $m$, between signatures $p_i$ and $p_j$ is given by summing the residuals as the two arrays are convolved over one another, described by \autoref{equ:ok_spec}.  

\begin{equation}
    m(p_i,p_j) = \frac{1}{dim} \sum_{dim} \frac{1}{|\kappa|} \sum_{\in \kappa} G \big[s - \sum^{\eta}_{n} \big[ p_i(n) - p_j(n) \big] \big]
    \label{equ:ok_spec}
\end{equation}

Where $\kappa$ is the set of all possible points of convolution between $p_i$ and $p_j$ such that $\kappa = 1:g:\tau$, where $g$ is a variable to determine the resolution of convolution, $\tau = ||p_i| - |p_j|| + k + 1$, and $k$ determines the permissible index offset between $p_i$ and $p_j$ during convolution. $n$ is the index of data points stored in $p_{i,j}$, and $\eta$ is the total number of overlaying data points at a given point of convolution. $G$ is a function such that $G(x) = 0$ for $x < 0$, which allows variable $s$ to act as a threshold such that a pair of signatures with residuals greater than $s$ at a given convolution point are considered not to match at all. $dim$ refers to the number of dimensions of $p_{i,j}$. 

Each robot records an independent sliding behavioural window $W$, containing its 300 most recent values of $v,\omega,\Delta P$ ($W_M$) and $\gamma$ ($W_S$) -- i.e., the most recent 50 seconds of behavioural data. After every 50 seconds of simulated time, each robot updates the population scores of every behavioural signature in its repertoires $X_M$ and $X_S$ according to \autoref{equ:ok}.

\begin{equation}
\dot{x_i} = m(p_i, W_0)\cdot \big[1 + k_3 max( m(p_i, Y)) \big] - k_1 \sum_{j=1}^{N-1} m(p_i, W_j)  - k_2 
\label{equ:ok}
\end{equation}

Where $x_i$ is the population score of the $i^{th}$ behavioural signature in a given robot's repertoire $X_{M,S}$. $W_0$ is a robot's own behavioural window, and $W_j$ is the equivalent window for the $j^{th}$ robot in the swarm. $k_{1-3}$ are tuning parameters. $Y$ is a repertoire containing behavioural signatures that have been identified and labelled as faulty \textit{a priori}. In this work, $Y_{M,S}$ contain 101 and 93 labelled behavioural signatures, respectively, which are kept constant and shared across the SRS.

The fault detection model can be described thus; the population score $x_i$ of a behavioural signature $p_i$ contained in $X$ is stimulated according to how closely it matches with a robots own recent behavioural window $W_0$ -- i.e, a behavioural signature must be exhibited persistently by a robot for its population score to increase. Population score $x_i$ is further stimulated if $p_i$ closely matches with a learned faulty behavioural signature $p_j$ contained in $Y$. Population score $x_i$ is also suppressed according to how closely it matches with the behavioural windows of other robots in the swarm -- i.e, a behavioural signature exhibited by a majority of robots has its population score suppressed. A behavioural signature is considered faulty if its population score $x_i > 1$, while a behavioural signature with population score $x_i < 0$ is removed from $X$. Each robot runs its own instance of the model, summarised in \autoref{algAAPD}.

A variety of parameter values are used for \autoref{equ:ok_spec} and \autoref{equ:ok} according to the stage of \autoref{algAAPD} and whether it is operating on behavioural signatures contained within $X_M$ or $X_S$. These are provided in \autoref{table:param_values}.

\begin{algorithm} [t]
\caption{Fault Detection Algorithm}\label{algAAPD}
\begin{algorithmic}[1]
\While {Running}
\If{time elapsed since last captured signature $p_{i-1}$ is $\geq 5 sec$} capture new behavioural signature $p_i$.
\EndIf
\If {No $p_j$ exists in $X$ such that $m(p_i,p_j) > 1.5$ according to \autoref{equ:ok_spec}} $p_i \cup X$ and assign population $x_i = 0$.
\EndIf
\If{time elapsed since last computation of population values is $ \geq 50sec$} compute new population values for every member of $X$ according to \autoref{equ:ok}.
\EndIf
\If{$p_i$ in $X$ has $x_i > 1$} detect $p_i$ as faulty signature.
\ElsIf{$p_i$ in $X$ has $x_i < 0$} remove $p_i$ from $X$
\EndIf
\EndWhile
\end{algorithmic}
\end{algorithm}

\begin{table}
\begin{center}
\begin{tabular}{ |l|c|c|c|c|c|c|c|}
\hline
Process & Computation & $s$ & $g$ & $k$ & $k_1$ & $k_2$ & $k_3$ \\
\hline
$p_i \cup X$ & \autoref{equ:ok_spec} & 1.5 & 1 & 10 & - & - & - \\
$m(p_i,W)$ (mot.) & \autoref{equ:ok_spec} & 4 & 5 & 0 & - & - & - \\
$m(p_i,Y)$ (mot.) & \autoref{equ:ok_spec} & 1.5 & 1 & 10 & - & - & - \\
$\dot{x_i}$ (mot.) & \autoref{equ:ok} & - & - & - & 0.24 & 0.3 & 1.2 \\
$m(p_i,W)$ (sen.) & \autoref{equ:ok_spec} & 5 & 5 & 0 & - & - & - \\
$m(p_i,Y)$ (sen.) & \autoref{equ:ok_spec} & 3.3 & 1 & 10 & - & - & - \\
$\dot{x_i}$ (sen.) & \autoref{equ:ok} & - & - & - & 0.18 & 0.3 & 1.2 \\
\hline

\end{tabular}
\end{center}
\caption{The parameter selections for \autoref{equ:ok} and \autoref{equ:ok_spec} used at different stages of the fault detection model described in \autoref{algAAPD}.}
\label{table:param_values}
\end{table}

The model operates on signatures contained in $X_S$ at all times. However, a minimum of 5 robots are required for reliable performance, and so the model will only operate on signatures in $X_M$ for a robot performing the LPF algorithm if it and at least 4 other robots are free to move simultaneously. 

\subsection*{Fault Resolution}

A predictive fault resolution, $T_P$, is compared with a reactive fault resolution, $T_R$. 

\subsubsection*{$T_P$}
Faulty robots return themselves to the robot base where they are either replaced or redeployed, having been assumed to receive any necessary maintenance work. Resources carried back to the base are counted towards performance. $T_P$ relies on the ability to detect a fault while the afflicted robot is still sufficiently operational to make the return journey to base -- i.e. before the fault has chance to manifest as failure. If the robot fails to return to base, it becomes stranded and no replacement is deployed.

\subsubsection*{$T_R$} Faulty robots are shutdown. If the robot was not within the base area at the time of detection, it becomes an inanimate object and any resource it was carrying is not counted towards performance. A robot that is detected as faulty while in the base is considered to be reachable and is therefore collected and removed, along with any resource it was carrying. In each case a replacement is spawned at the robot base. This is the basic resolution that satisfies the vulnerability of a SRS to partial failures \cite{winfield2006safety}, and which has been implemented in previous research \cite{khadidos2015exogenous} \cite{strobel2023robot}.

\section{Experiments, Results and Discussion}

A baseline performance is established for swarms of size $N$ = 5, 10, and 20 robots performing the GPF-algorithm and the LPF-algorithm in empty and constrained environments, as well as the effects of faults on individual robot and overall performance in each of these scenarios where unchecked sensor and motor faults afflict sub-populations of 20\%, 40\%, and 60\% of the swarm size. The median number of resources collected by each robot in 15 minutes of simulated time is selected as the performance metric, since this value will vary between robots according to robot status, environment and behaviour type, as well as random probability. Afflicted robots are given a 33\% probability of $d_{l,r,S}$ decrementing by 0.01 per second of simulated time, meaning that failures have typically fully manifested after 5 minutes of simulated time. 10 experimental replicates are performed. Results are displayed in \autoref{table:Bases}. 

\autoref{table:Bases} shows that, where all robots are normally operational, a constrained environment results in fewer resources being collected in the same time in each case. The ideal swarm size varies according to the scenario. The LPF-algorithm requires a greater number of robots to function. $N$ = 5 is generally insufficient, as robots are stretched to the limits of their communication range, becoming stuck, while $N$ = 10 gives reduced performance compared with the GPF-algorithm in the constrained environment. Otherwise, however, the LPF and GPF algorithms are shown to perform competitively. 

\autoref{table:Bases} shows that, in open environments, motor failures can severely impede the afflicted robot while having little to no impact on others -- demonstrating characteristic SRS robustness. The impact of sensor failure during the GPF-algorithm is minor, since the sensing range must drop to near-zero before the afflicted robot becomes unable to avoid collisions or come across the resource nests by chance. However, when movement is constrained, faulty robots begin to have a more pronounced effect on non-faulty robots. In particular, robots that become stuck in corridors can obstruct other robots from passing (as shown in \autoref{fig:setup}D). Robots performing the LPF-algorithm with one or more chained dependents can also cause their dependents to become temporarily or permanently stranded when their sensor range drops too low. The circumstances of robot failure can also have net positive effects. For example, traffic around resource nests or in corridors can be reduced if robots fail outside these areas, improving the performance of normal robots. Overall, the swarm demonstrates fault tolerance except in cases where the number of functioning robots is brought below a required threshold, or where other robots in the swarm are dependent on a failed robot, as is expected \cite{csahin2004swarm,bjerknes2013fault}.

\begin{table*}[!t]
\begin{center}
\resizebox{\textwidth}{!}{\begin{tabular}{|l|c|c|c|c|c|c|c|c|c|c|c|c|c|c|c|c|c|c|c|c|c|c|c|c|}
\hline
\multicolumn{1}{|c|}{} & \multicolumn{12}{c|}{Open Environment} & \multicolumn{12}{c|}{Constrained Environment}\\
\cline{2-25}
\multicolumn{1}{|c|}{} & \multicolumn{6}{c|}{GPF} & \multicolumn{6}{c|}{LPF} & \multicolumn{6}{c|}{GPF} & \multicolumn{6}{c|}{LPF} \\
\cline{2-25}
\multicolumn{1}{|c|}{} & \multicolumn{2}{c|}{N = 5} & \multicolumn{2}{c|}{N = 10} & \multicolumn{2}{c|}{N = 20} & \multicolumn{2}{c|}{N = 5} & \multicolumn{2}{c|}{N = 10} & \multicolumn{2}{c|}{N = 20} & \multicolumn{2}{c|}{N = 5} & \multicolumn{2}{c|}{N = 10} & \multicolumn{2}{c|}{N = 20} & \multicolumn{2}{c|}{N = 5} & \multicolumn{2}{c|}{N = 10} & \multicolumn{2}{c|}{N = 20}  \\
\cline{2-25}
\multicolumn{1}{|c|}{Fault$^*$\%} & $R$ & $R^*$ & $R$ & $R^*$
 & $R$ & $R^*$ & $R$ & $R^*$ & $R$ & $R^*$ & $R$ & $R^*$
 & $R$ & $R^*$ & $R$ & $R^*$  & $R$ & $R^*$ & $R$ & $R^*$  & $R$ & $R^*$ & $R$ & $R^*$\\
\hline

None & 10 & -- & 10 & -- & 8 & -- & 2 & -- & 10 & -- & 8 & -- & 6 & -- & 6 & -- & 5 & -- & 0 & -- & 4 & -- & 5 & -- \\
\hline
M, 20\% & 11 & 1 & 10 & 1 & 9 & 1 & 3 & 1 & 10 & 1 & 9 & 1 & 7 & 1 & 6 & 1 & 6 & 1 & 0 & 0 & 0 & 0 & 5 & 0 \\
M, 40\% & 12 & 1 & 9 & 1 & 10 & 1 & 1 & 1 & 10 & 1 & 9 & 1 & 8 & 1 & 8 & 1 & 6 & 1 & 0 & 0 & 1 & 0 & 5 & 1 \\
M, 60\% & 12 & 1 & 11 & 1 & 9 & 1 & 1 & 1 & 9 & 1 & 9 & 1 & 7 & 1 & 7 & 1 & 5 & 1 & 0 & 0 & 1 & 1 & 4 & 1 \\
\hline
S, 20\% & 11 & 7.5 & 11 & 6.5 & 9 & 5.5 & 1 & 1 & 9 & 5 & 9 & 6 & 7 & 6 & 7 & 5 & 5 & 3 & 0 & 0 & 2 & 0 & 4 & 3 \\
S, 40\% & 12 & 7 & 11 & 7 & 9 & 6 & 2 & 1 & 8 & 5 & 9 & 5 & 7 & 6 & 5 & 4 & 5 & 4 & 0 & 0 & 0 & 2 & 3 & 2 \\
S, 60\% & 12 & 9 & 11 & 6 & 9 & 5 & 2 & 2 & 5 & 5 & 8 & 5 & 7 & 6 & 7 & 5 & 6 & 3 & 0 & 0 & 0 & 0.5 & 2 & 1 \\
\hline

\end{tabular}}
\end{center}
\caption{The median number of resources collected per robot in 15 minutes of simulated time in each combination of algorithm, environment, and swarm size. Performances of normal and faulty robots are given separately, where between 0-60\% of the swarm suffers from motor or sensor faults, denoted `M' or `S', respectively. Normal and faulty robots are denoted `R' and `R$^*$', respectively.}
\label{table:Bases}
\end{table*}

\subsection*{Reactive Vs. Predictive Resolution}

\begin{figure*}[!htb]
    \includegraphics[width=\textwidth]{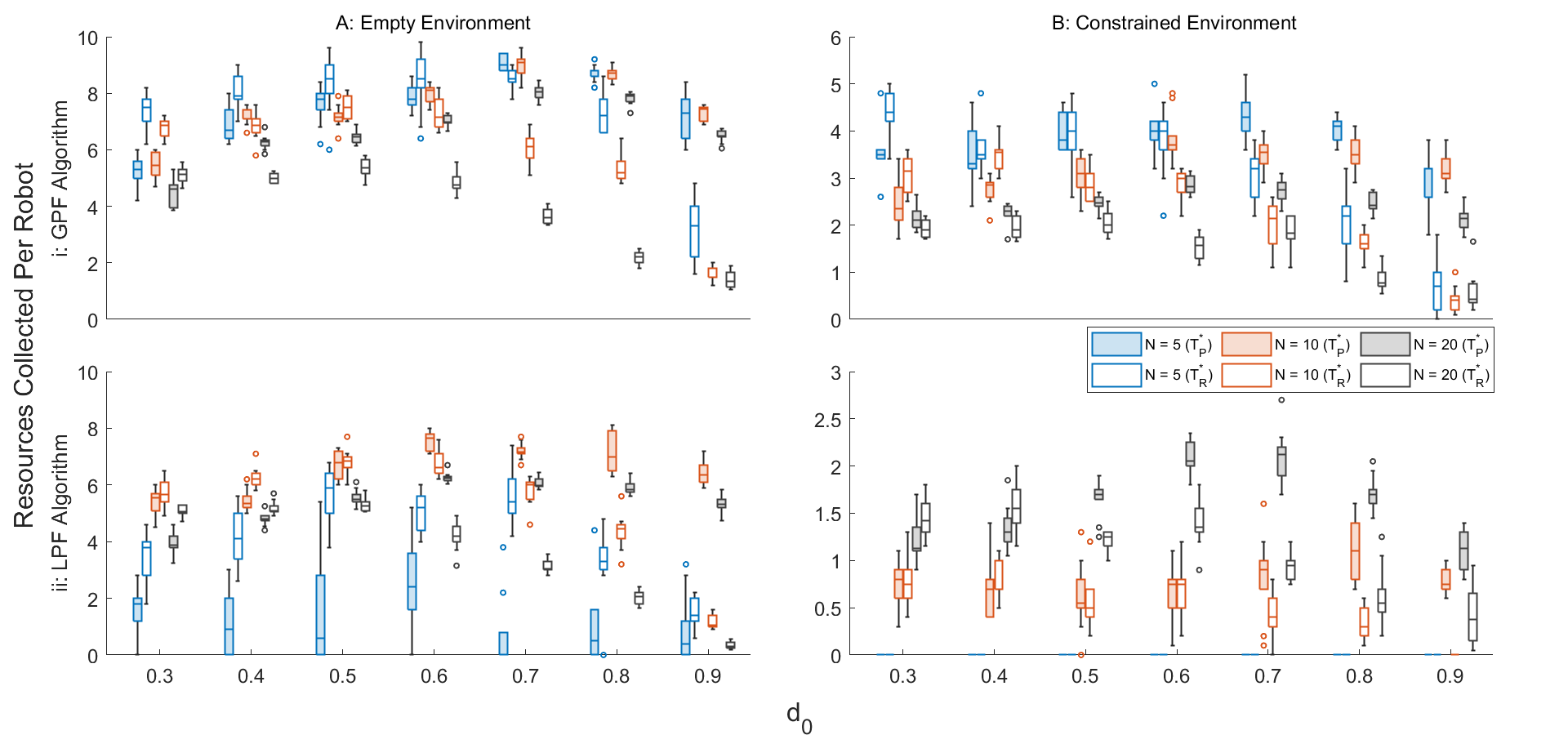}
    \caption{The median resources collected by each robot in 15 minutes of simulated time for every combination of algorithm, environment, and swarm size. A comparison is shown for predictive ($T_P^*$) 
 and reactive ($T_R^*$) fault resolutions, displayed as filled in or white bars, respectively. Fault resolutions are initiated for robots with any $d_{l,r,S} < d_0$.} 
    \label{fig:PFDDRAnalysis}  
\end{figure*}

In the following experiments, $d_{l,r,S}$ are each given random independent probabilities between 1-15\% of decreasing by 0.01 per second, reflecting that each robot has its own MTBF. In order to establish the ideal point of fault detection, a robot is automatically detected as faulty when any value $d_{l,r,S}$ drops below threshold $d_0$. The differences in performance where faulty robots are resolved via the predictive maintenance ($T_P^*$) and reactive shutdown ($T_R^*$) approaches are then observed for varying $d_0$. The use of asterisks is to differentiate these experiments, which use an ideal fault detection mechanism, from subsequent experiments using the fault detection model described by \autoref{algAAPD}. Results are displayed in \autoref{fig:PFDDRAnalysis}.

\autoref{fig:PFDDRAnalysis} shows that, for any combination of algorithm and environment, a consistent trend can be observed between $d_0$ and performance according to whether the system implements $T_P^*$ or $T_R^*$. In the case of $T_P^*$, there is a defined optimum region, typically in the range $0.6 < d_0 < 0.8$. For $d_0 < 0.6$, the time that robots spend operating with degraded capabilities reduces performance and can increase the difficulty in making the return journey to the base, and thus the length of interruption to foraging activity. For $d_0 < 0.8$, the increased frequency of detected faults means that robots similarly have their foraging activity interrupted for a higher proportion of experimental time. In the case of $T_R^*$, there is a less defined optimum, with median performance tending to plateau for $0.3 < d_0 < 0.6$, and then decreasing for $d_0 > 0.6$. The reason for this is that shut-down robots obstruct the normally functioning swarm. In the constrained environment, this can result in the complete prevention of foraging, as the corridors become impassable. This effect is not seen to the same degree in \autoref{table:Bases}, since faulty robots are not replaced, and is a direct consequence of $T_R$. Disruption can also be observed in the open environment because of the tendency of failed robots to form clusters. A failed robot acts as an obstacle to a normally operating robot, which must avoid it upon encounter or, if the robot has already identified a goal on the other side of the obstacle, must navigate around it. The process of navigating the obstacle increases the amount of time the operational robot spends in proximity to the failed robot, and therefore the likelihood that it, too, will fail in close proximity -- increasing the size of the obstacle and thereby making it subsequently harder to navigate. This creates a feedback loop in which each failed robot contributes to a decrease in the reliability of SRS performance, even when it is replaced. 

It was expected that the additional time taken to implement $T_P^*$ would put it at a disadvantage to $T_R^*$ in scenarios where the SRS was expected to be most robust to failure (e.g. open environments, GPF algorithm). However, \autoref{fig:PFDDRAnalysis} reveals that, in addition to several scenarios in which $T_P^*$ outperforms $T_R^*$, there are almost no scenarios in which $T_P^*$ cannot at least give a competitive performance with $T_R^*$. It should also be noted that, where $T_P^*$ and $T_R^*$ are competitive, $T_P^*$ offers an innate real-world advantage over $T_R^*$ in its potential to repair and reuse hardware resources. This result is significant as it underscores the value of a predictive approach to fault tolerance in SRS for the first time, and challenges the utility of reactive approaches in scenarios where robots cannot be physically retrieved. The only notable exception to this trend is in the case of $N = 5$ robots performing the LPF algorithm in the open environment. This comes from the fact that 5 robots is too few to effectively implement the LPF algorithm (as shown in \autoref{table:Bases}), resulting in the robots becoming stretched to their limits and stuck. The replacement of a robot allows it to benefit from the existing network coverage provided by the remaining robots, often meaning it is able to collect resources successfully for a period until it, too, becomes stuck. This highlights the need for intelligent path planning of the return to base journey in scenarios where the relative positioning of robots is critical.

\begin{figure}[!htb]
  \centering  
    \includegraphics[width=0.48\textwidth]{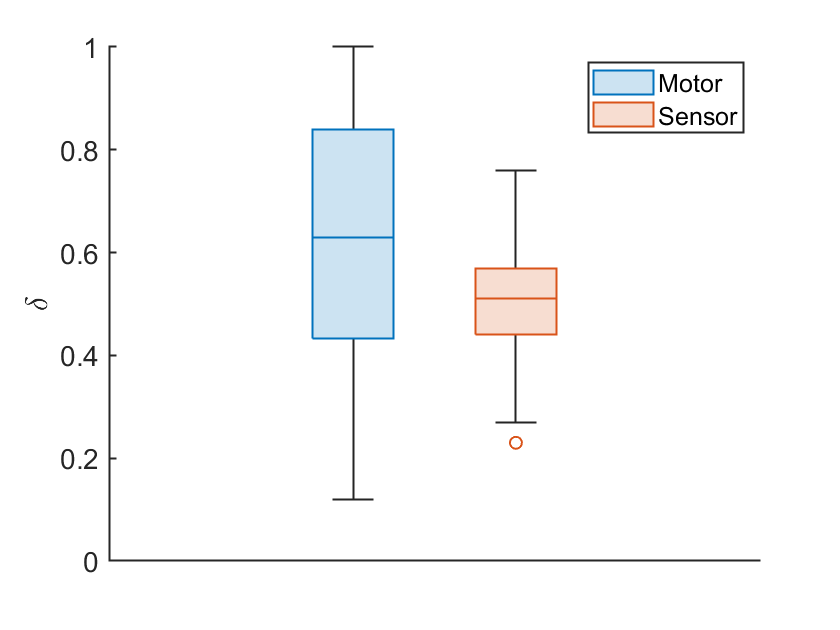}
    \caption{The values of $\delta$ for detections of motor and sensor faults made by \autoref{algAAPD} on $N = 10$ robots performing the GPF algorithm in the open environment for 15 minutes of simulated time.} 
    \label{fig:AAPD_Dets}  
\end{figure}

The final set of experiments implements the fault detection model described in \autoref{algAAPD} as a means of detecting faulty robots. The performance of the model is characterised in \autoref{fig:AAPD_Dets}, which plots the $\delta$ values for each fault detected by the model when deployed on $N = 10$ robots performing the GPF algorithm in the open environment. The value $\delta$ indicates the value $d_S$ of a given robot at the moment \autoref{algAAPD} detects a sensor fault or the value of $d_{l,r}$, whichever is lowest, at the moment it detects a motor fault. Each robot is initialised with a random probability between 1-15\% that $d_{l,r}$ and $d_s$ will decrease by 0.01 per second of simulated time. In the case of motor faults, the AAPD detects faulty robots with a median $\delta = 0.63$, within the optimum range for $d_0$ in each scenario highlighted in \autoref{fig:PFDDRAnalysis}, albeit with a larger than desirable interquartile range. In the case of sensor faults, the AAPD detects faulty robots with a median $\delta = 0.52$. This is lower than desired, but sensor faults are harder to detect reliably since it is not always possible to detect drops in range according to the $\gamma$ value produced. Sensor faults are also generally less disruptive than motor faults (see \autoref{table:Bases}), and so the lower than ideal $\delta$ is not expected to be as punishing on performance. 

\begin{figure*}[!htb]
    \includegraphics[width=\textwidth]{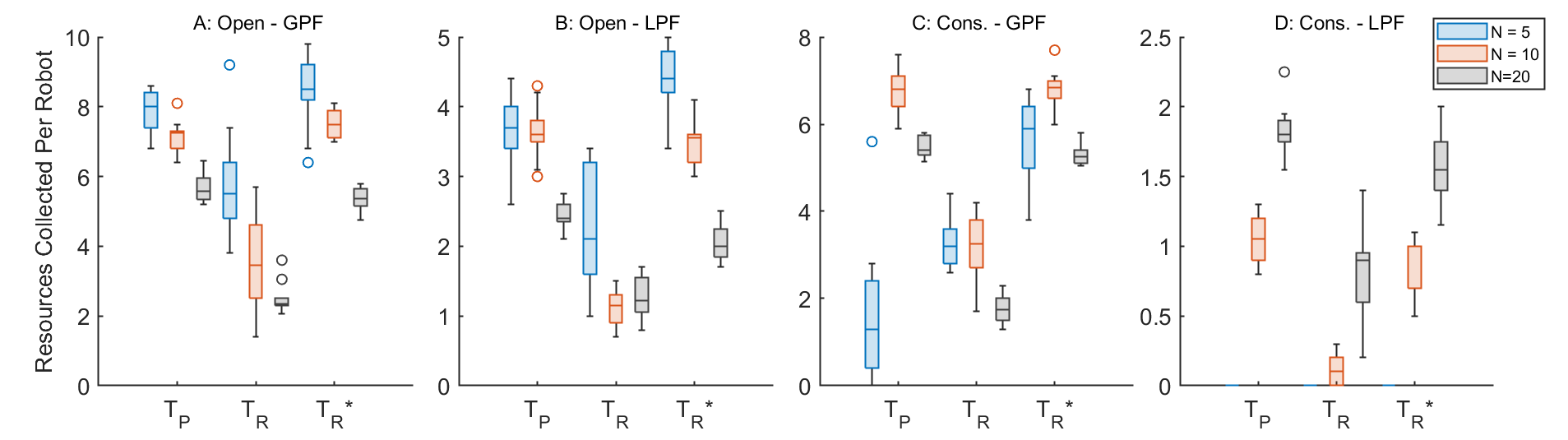}
    \caption{The median resources collected in 15 minutes of simulated time in each combination of algorithm, environment, and swarm size. A comparison is shown for predictive ($T_P$) 
 and reactive ($T_R$) fault resolutions initiated when a fault is detected by \autoref{algAAPD}, as well as the highest performing instance of $T_R^*$ taken from the corresponding scenario in \autoref{fig:PFDDRAnalysis}.}
    \label{fig:AAPDAnalysis}  
\end{figure*}

Overall SRS performance when \autoref{algAAPD} is implemented is shown in \autoref{fig:AAPDAnalysis}, which plots a comparison of resolutions $T_P$ and $T_R$ for each combination of swarm size, algorithm, and arena type. Each robot is initialised with random independent probabilities between 1-15\% that $d_{l,r}$ and $d_S$ will decrease by 0.01 per second of simulated time. Also included is the performance for $T_R^*$ with the optimum $d_0$ value taken from \autoref{fig:PFDDRAnalysis}. This is done to compare the performance of predictive resolution $T_P$ using \autoref{algAAPD} with the theoretic optimum performance for reactive approach $T_R^*$.

\autoref{fig:AAPDAnalysis} shows that, where the AAPD model is used to detect faults, predictive resolution $T_P$ substantially outperforms reactive resolution $T_R$ in all cases with the exception of $N$ = 5 robots performing the LPF foraging algorithm, which is discussed earlier. This is expected, since \autoref{fig:AAPDAnalysis} typically detects faults with $\delta$ values that \autoref{fig:PFDDRAnalysis} reveals to be suboptimal when implementing $T_R^*$. The performance of $T_R$ could, in theory, be improved by modifying \autoref{fig:AAPDAnalysis} parameters to tolerate faults at lower values of $\delta$. However, \autoref{fig:AAPDAnalysis} also shows that implementing $T_P$ with \autoref{fig:AAPDAnalysis} gives a generally competitive performance against optimum $T_R^*$ -- typically to within a few percent, but as low as -16\% and as high as +50\%. This result further supports the adoption of predictive approaches to fault tolerance in SRS. For ease of reading the proportional difference in median performances, plotted in \autoref{fig:AAPDAnalysis} of $T_P$ against $T_R$ and $T_R^*$, are given in \autoref{table:AAPD}.

\begin{table}[!t]
\begin{center}
\resizebox{0.48\textwidth}{!}{\begin{tabular}{|l|c|c|c|c|c|c|c|c|}
\hline
\multicolumn{1}{|c|}{} & \multicolumn{4}{c|}{Open Environment} & \multicolumn{4}{c|}{Constrained Environment}\\
\cline{2-9}
\multicolumn{1}{|c|}{} & \multicolumn{2}{c|}{GPF} & \multicolumn{2}{c|}{LPF} & \multicolumn{2}{c|}{GPF} & \multicolumn{2}{c|}{LPF} \\
\cline{2-9}

\multicolumn{1}{|c|}{} & $T_R$ & $T_R^*$ & $T_R$ & $T_R^*$ & $T_R$ & $T_R^*$ & $T_R$ & $T_R^*$ \\
\hline
\multicolumn{1}{|c|}{N = 5} & +64 & -6 & -60 & -78 & +76 & -16 & -- & -- \\
\hline
\multicolumn{1}{|c|}{N = 10} & +110 & -3 & +109 & -1 & +65 & -1 & +950 & +50 \\
\hline
\multicolumn{1}{|c|}{N = 20} & +137 & +6 & +209 & +3 & +96 & +20 & +100 & +16 \\
\hline

\hline
\end{tabular}}
\end{center}
\caption{The proportional difference (as a percentage) in median performance achieved by $T_P$ when compared to $T_R$ and $T_R^*$, taken from \autoref{fig:AAPDAnalysis}.}
\label{table:AAPD}
\end{table}

\section{Conclusion and Future Work}

This paper highlights that the study of faults arising from gradual sensor and actuator degradation in robots has been absent from fault tolerant swarm literature until now, despite being a very common cause of real world failure. A variety of generalisable swarm foraging scenarios, in which every robot has a variable MTBF, demonstrate that motor and sensor failures can cause large reductions in swarm foraging performance for certain combinations of behaviour, environment, and swarm size. Where previous research has highlighted the vulnerability of robot swarms to partial failures in individuals, this study underscores the vulnerabilities associated with allowing robots to reach a point of failure if they are obstructive and cannot be physically removed from an environment. 

As an autonomous solution, a predictive approach to swarm fault tolerance is proposed in which faults are detected with enough time to allow the at-risk robot to autonomously return itself to a safe area to receive maintenance or be replaced. The ability of at-risk robots to autonomously return themselves to a safe location is critical, since robots are not typically able to repair themselves or others in the field, and thus faults must be detected before their manifestation as complete failures. The prevention of failures in swarm robots is paradoxical to traditionally held views that robot swarms should be robust to the loss of individuals. However, results show that detecting faults with enough time for the robot to remove itself from harms way results in competitive or improved performance when compared to allowing robots to fail in the field in nearly all cases tested. This is verified with a theoretic ideal fault detection mechanism, as well as with a novel swarm fault detection algorithm. In addition to the empirical evidence in support of predictive fault tolerance in swarms, it comes with an innate real-world advantage insofar that it allows for the conservation of hardware resources by repairing and reusing robots instead of simply abandoning them. Overall, the results show that a predictive approach to fault tolerance allows a swarm to sustain its own autonomy for longer periods of time, and potentially allows the swarm to operate in environments where susceptibility to failures could have made it otherwise unsuitable. This represents a departure from traditional approaches to swarm fault tolerance, and is therefore an important contribution to the literature that highlights new areas for exploration in future research. 

A key takeaway from the results is the importance of detecting faults at the right moment so that robots are not allowed to reach states where their autonomy is put at risk, while also ensuring that operation is not needlessly interrupted. It is critically important that a robot detected as faulty is able to reach a safe location, and in real world scenarios this will influence the ideal point of detection -- e.g. the greater the distance to the nearest safe location, the lower the level of tolerable degradation can be. Other influencing factors include the spatial limitations imposed by the environment and other robots in some scenarios. Future work will therefore focus on improving the reliability of fault detection to within an optimal range of degradation, as well as incorporating intelligent online path planning into fault resolution in order to achieve robust closed-loop FDDR.

\bibliographystyle{ieeetr}
\bibliography{master1}

\vfill

\end{document}